\documentclass[sigconf]{acmart}

\copyrightyear{2024}
\acmYear{2024}
\setcopyright{rightsretained}
\acmConference[KDD '24]{Proceedings of the 30th ACM SIGKDD Conference on Knowledge Discovery and Data Mining}{August 25--29, 2024}{Barcelona, Spain}
\acmBooktitle{Proceedings of the 30th ACM SIGKDD Conference on Knowledge Discovery and Data Mining (KDD '24), August 25--29, 2024, Barcelona, Spain}\acmDOI{10.1145/3637528.3671717}
\acmISBN{979-8-4007-0490-1/24/08}

\makeatletter
\gdef\@copyrightpermission{
  \begin{minipage}{0.3\columnwidth}
   \href{https://creativecommons.org/licenses/by/4.0/}{\includegraphics[width=0.90\textwidth]{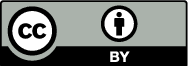}}
  \end{minipage}\hfill
  \begin{minipage}{0.7\columnwidth}
   \href{https://creativecommons.org/licenses/by/4.0/}{This work is licensed under a Creative Commons Attribution International 4.0 License.}
  \end{minipage}
  \vspace{5pt}
}
\makeatother

\usepackage{booktabs}
\usepackage{boldline} 
\usepackage{multirow}
\usepackage{colortbl}
\usepackage{xcolor}
\usepackage{array}
\usepackage{cleveref}
\usepackage{algorithm}
\usepackage{algorithmic}
\usepackage{balance}

\allowdisplaybreaks

\begin{document}

\title{Federated Graph Learning with Structure Proxy Alignment}


\author{Xingbo Fu}
\affiliation{%
\institution{University of Virginia}
\city{Charlottesville}
\state{Virginia}
\country{USA}}
\email{xf3av@virginia.edu}

\author{Zihan Chen}
\affiliation{%
\institution{University of Virginia}
\city{Charlottesville}
\state{Virginia}
\country{USA}}
\email{brf3rx@virginia.edu}

\author{Binchi Zhang}
\affiliation{%
\institution{University of Virginia}
\city{Charlottesville}
\state{Virginia}
\country{USA}}
\email{epb6gw@virginia.edu}

\author{Chen Chen}
\affiliation{%
\institution{University of Virginia}
\city{Charlottesville}
\state{Virginia}
\country{USA}}
\email{zrh6du@virginia.edu}

\author{Jundong Li}
\affiliation{%
\institution{University of Virginia}
\city{Charlottesville}
\state{Virginia}
\country{USA}}
\email{jundong@virginia.edu}

\renewcommand{\shortauthors}{Xingbo Fu, Zihan Chen, Binchi Zhang, Chen Chen, \& Jundong Li}
\begin{abstract}
Federated Graph Learning (FGL) aims to learn graph learning models over graph data distributed in multiple data owners, which has been applied in various applications such as social recommendation and financial fraud detection. Inherited from generic Federated Learning (FL), FGL similarly has the data heterogeneity issue where the label distribution may vary significantly for distributed graph data across clients. For instance, a client can have the majority of nodes from a class, while another client may have only a few nodes from the same class. This issue results in divergent local objectives and impairs FGL convergence for node-level tasks, especially for node classification. Moreover, FGL also encounters a unique challenge for the node classification task: the nodes from a minority class in a client are more likely to have biased neighboring information, which prevents FGL from learning expressive node embeddings with Graph Neural Networks (GNNs). To grapple with the challenge, we propose FedSpray, a novel FGL framework that learns local class-wise structure proxies in the latent space and aligns them to obtain global structure proxies in the server. Our goal is to obtain the aligned structure proxies that can serve as reliable, unbiased neighboring information for node classification. To achieve this, FedSpray trains a global feature-structure encoder and generates unbiased soft targets with structure proxies to regularize local training of GNN models in a personalized way. We conduct extensive experiments over four datasets, and experiment results validate the superiority of FedSpray compared with other baselines. 
Our code is available at https://github.com/xbfu/FedSpray.
\end{abstract}



\maketitle

\section{Introduction}
\begin{figure}[t]
\setlength {\belowcaptionskip} {-0.4cm}
\centering
\includegraphics[width=\linewidth]{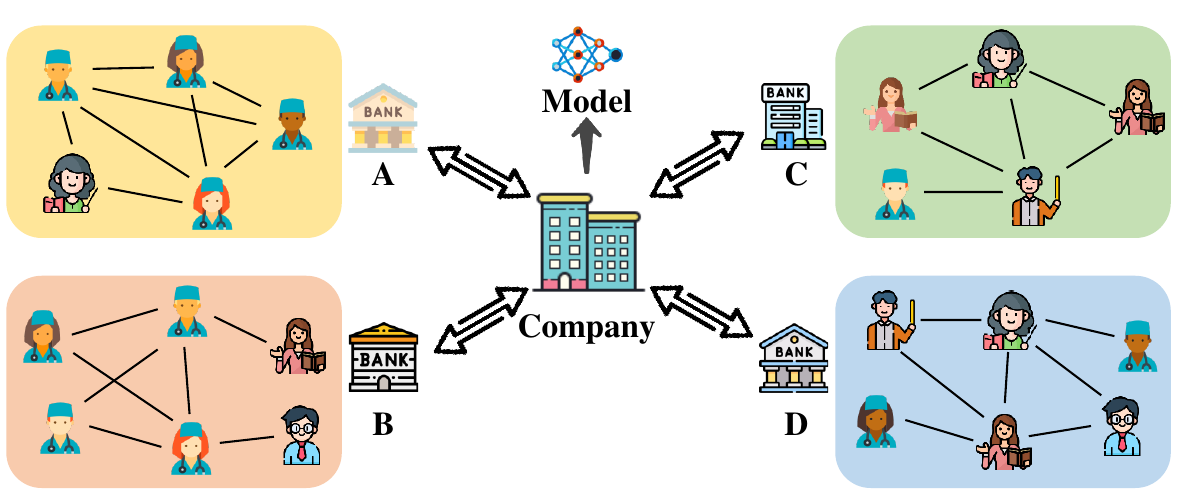}
\caption{An example of a financial system including four banks. The four banks aim to jointly train a model for predicting a customer's occupation (i.e., \textit{Doctor} or \textit{Teacher}) orchestrated by a third-party company over their local data while keeping their private data locally.}
\label{fig:example}
\end{figure}

Graph Neural Networks (GNNs) \cite{wu2020gnnsurvey1} are a prominent approach for learning expressive representations from graph-structured data. Typically, GNNs follow a message-passing mechanism, where the embedding of each node is computed by aggregating attribute information from its neighbors \cite{kipf2016gcn,hamilton2017graphsage,wu2019sgc}.
Thanks to their powerful capacity for jointly embedding attribute and graph structure information, GNNs have been widely adopted in a wide variety of applications, such as node classification \cite{fu2023spatial,he2022variational} and link prediction \cite{cai2021link_pred1,daza2021link_pred2}.
The existing GNNs are mostly trained in a centralized manner where graph data is collected on a single machine before training. In the real world, however, a large number of graph data is generated by multiple data owners. These graph data cannot be assembled for training due to privacy concerns and commercial competitions \cite{wang2024safety}, which prevents the traditional centralized manner from training powerful GNNs. 
Taking a financial system with four banks in Figure~\ref{fig:example} as an example, each bank in the system has its local customer dataset and transactions between customers. As we take the customers in a bank as nodes and transactions between them as edges, the bank's local data can naturally form a graph. These banks aim to jointly train a GNN model for classification tasks, such as predicting a customer's occupation (i.e., \textit{Doctor} or \textit{Teacher}) without sharing their local data with each other.

Federated Learning (FL) \cite{mcmahan2017fl} is a prevalent distributed learning scheme that enables multiple data owners (i.e., clients) to collaboratively train machine learning models under the coordination of a central server without sharing their private data. 
One critical challenge in FL is data heterogeneity, where data samples are not independent and identically distributed (i.e., non-IID) across the clients. For instance, assume that Bank A in Figure~\ref{fig:example} locates in a community adjacent to a hospital. Then most customers in Bank A are therefore likely to be labeled as \textit{Doctor} while only a few customers are from other occupations (e.g., \textit{Teacher}). In contrast, Bank C adjoining a school has customers labeled mostly as \textit{Teacher} and only a few as \textit{Doctor}. Typically, the nodes from a class that claims the very large proportion of the overall data in a client are the \textit{majority nodes} (e.g., \textit{Doctor} in Bank A) while \textit{minority nodes} (e.g., \textit{Teacher} in Bank A) account for much fewer samples. The data heterogeneity issue results in divergent local objectives on the clients and consequently impairs the performance of FL \cite{karimireddy2020scaffold}. A number of approaches have been proposed to address this issue, to name a few \cite{li2020fedprox,wu2024breaking, wang2023federated}.

When we train GNNs over distributed graph data in a federated manner, however, the data heterogeneity issue can get much more severe. This results from a unique challenge in Federated Graph Learning (FGL) \cite{fu2022federated}: 
\textbf{the high heterophily of minority nodes}, i.e., their neighbors are mostly from other classes \cite{song2022tam}. A majority node in a client (e.g., \textit{Teacher} in Bank D) can benefit from the message-passing mechanism and obtain an expressive embedding as its neighbors are probably from the same class. On the contrary, a minority node in another client (e.g., \textit{Teacher} in Bank A) may obtain biased information from its neighbors when they are from other classes (e.g., \textit{Doctor} in Bank A). 
In FGL, this challenge is usually entangled with the data heterogeneity issue. As a result, the minority nodes will finally get underrepresented embeddings given adverse neighboring information and be more likely to be predicted as the major class, which results in unsatisfactory performance. Although a few studies have investigated the data heterogeneity issue about graph structures in FGL \cite{xie2021gcfl, tan2023fedstar}, they did not fathom the divergent impact of neighboring information across clients for node classification.

To tackle the aforementioned challenges in FGL, we propose FedSpray, a novel FGL framework with structure proxy alignment in this study. The goal of FedSpray is to learn personalized GNN models for each client while avoiding underrepresented embeddings of the minority nodes in each client caused by their adverse neighboring information in FGL. To achieve this goal, we first introduce global class-wise structure proxies \cite{dong2023reliant} which aim to provide nodes with informative, unbiased neighboring information, especially for those from the minority classes in each client. Moreover, FedSpray learns a global feature-structure encoder to obtain reliable soft targets that only depend on node features and aligned structure proxies. Then, FedSpray uses the soft targets to regularize local training of personalized GNN models via knowledge distillation \cite{hinton2015kd}. We conduct extensive experiments over five graph datasets, and experimental results corroborate the effectiveness of the proposed FedSpray compared with other baselines.

We summarize the main contributions of this study as follows.
\begin{itemize}
    \item \textbf{Problem Formulation.} We formulate and make an initial investigation on a unique issue of unfavorable neighboring information for minority nodes in FGL.
    \item \textbf{Algorithmic Design.} We propose a novel framework FedSpray to tackle the above problem in FGL. FedSpray aims to learn unbiased soft targets by a global feature-structure encoder with aligned class-wise structure proxies which provide informative, unbiased neighboring information for nodes and guide local training of personalized GNN models.
    \item \textbf{Experimental Evaluation.} We conduct extensive experiments over four graph datasets to verify the effectiveness of the proposed FedSpray. The experimental results demonstrate that our FedSpray consistently outperforms the state-of-the-art baselines.
\end{itemize}

\section{Problem Formulation}
\subsection{Preliminaries}
\subsubsection{\textbf{Notations}}
We use bold uppercase letters (e.g., $\textbf{X}$) to represent matrices. For any matrix, e.g., $\textbf{X}$, we denote its $i$-th row vector as $\textbf{x}_i$. We use letters in calligraphy font (e.g., $\mathcal{V}$) to denote sets. $|\mathcal{V}|$ denotes the cardinality of set $\mathcal{V}$.

\subsubsection{\textbf{Graph Neural Networks}}
Let $\mathcal{G}=(\mathcal{V}, \mathcal{E}, \textbf{X})$ denote an undirected attributed graph, where $\mathcal{V}=\{v_1, v_2, \cdots, v_n\}$ is the set of $|\mathcal{V}|$ nodes, $\mathcal{E}$ is the edge set, and $\textbf{X}\in \mathbb{R}^{|\mathcal{V}|\times d_x}$ is the node feature matrix. $d_x$ is the number of node features. Given each node $v_i \in \mathcal{V}$, $\mathcal{N}(v_i)$ denotes the set of its neighbors.
The ground-truth label of each node $v_i \in \mathcal{V}$ can be denoted as a $d_c$-dimensional one-hot vector $\textbf{y}_i$ where $d_c$ is the number of classes. 
The node homophily \cite{yan2022two, luan2022revisiting} is defined as
\begin{equation}
   h_i = \frac{|\{v_j|v_j \in \mathcal{N}(v_i) \text{ and } \textbf{y}_j = \textbf{y}_i\}|}{|\mathcal{N}(v_i)|},
\end{equation}
where $|\mathcal{N}(v_i)|$ denotes the degree of node $v_i$.
Typically, an $L$-layer GNN model $f$ parameterized by $\theta$ maps each node to the outcome space via a message-passing mechanism \cite{kipf2016gcn, hamilton2017graphsage}. Specifically, each node $v_i$ aggregates information from its neighbors in the $l$-th layer of a GNN model by
\begin{equation} \label{gnn}
   \textbf{h}^l_i = f_l(\textbf{h}^{l-1}_i, \{\textbf{h}^{l-1}_j: v_j \in \mathcal{N}(v_i)\};\theta_l),
\end{equation}
where $\textbf{h}^l_i$ is the embedding of node $v_i$ after the $l$-th layer $f_l$, and $\theta_l$ is the parameters of the message-passing function in $f_l$. The raw feature of each node $v_i$ is used as the input layer, i.e., $\textbf{h}_i^0=\textbf{x}_i$. For the node classification task, the node embedding $\textbf{h}^L_i$ after the final layer is used to compute the predicted label distribution $\hat{\textbf{y}}_i=\text{Softmax}(\textbf{h}^L_i) \in \mathbb{R}^{d_p}$ by the softmax operator. 

\subsubsection{\textbf{Personalized FL}}
Given a set of $K$ clients, each client $k$ has its private dataset $\mathcal{D}^{(k)}=\{(\textbf{x}_i^{(k)}, \textbf{y}_i^{(k)})\}_{i=1}^{N^{(k)}}$, where $N^{(k)}$ is the number of samples in client $k$. The overall objective of the clients is
\begin{equation}
    \min\limits_{(\theta^{(1)},\theta^{(2)},\cdots,\theta^{(K)})}\sum_{k=1}^{K} \frac{N^{(k)}}{N}\mathcal{L}^{(k)}(\mathcal{D}^{(k)};\theta^{(k)}),
\end{equation}
where $\mathcal{L}^{(k)}(\theta^{(k)})$ is the local average loss (e.g., the cross-entropy loss) over local data in client $k$, and $N=\sum_{k=1}^{K}N^{(k)}$. Standard FL methods aim to learn a global model $\theta=\theta^{(1)}=\theta^{(2)}=\cdots=\theta^{(K)}$. As a representative method in FL, FedAvg \cite{mcmahan2017fl} performs local updates in each client and uploads local model parameters to a central server, where they are averaged by 
\begin{equation}
    \theta=\sum\limits_{k=1}^{K}\frac{N^{(k)}}{N} \theta^{(k)}
\end{equation}
during each round. However, a single global model may have poor performance due to the data heterogeneity issue in FL \cite{li2021ditto}. To remedy this, personalized FL \cite{tan2022pflsurvey1} allows a customized $\theta^{(k)}$ in each client $k$ with better performance on local data while still benefiting from collaborative training.

\subsection{Problem Setup}
Given a set of $K$ clients, each client $k$ owns a local graph $\mathcal{G}^{(k)}=(\mathcal{V}^{(k)}, \mathcal{E}^{(k)}, \textbf{X}^{(k)})$. For the labeled node set $\mathcal{V}_L^{(k)} \subset \mathcal{V}^{(k)}$ in client $k$, each node $v_i^{(k)} \in \mathcal{V}_L^{(k)}$ is associated with its label $\textbf{y}_i^{(k)}$.
The goal of these clients is to train personalized GNN models $f(\theta^{(k)})$ in each client $k$ for the node classification task while keeping their private graph data locally. Based on the aforementioned challenge and preliminary analysis, this study aims to enhance collaborative training by mitigating the impact of adverse neighboring information on node classification, especially for minority nodes.

\begin{table}[t]
\caption{The statistics of the majority class and other minority classes in 7 clients from the PubMed dataset. \textit{Majority} and \textit{Minority} represent the majority class and other minority classes, respectively.}
\begin{tabular}{c|c|cc|cc}
\hlineB{2}
\multirow{2}{*}{Client} & Majority & \multicolumn{2}{c|}{Num. of Nodes}  & \multicolumn{2}{c}{Avg. Homophily} \\ 
    & Class  & Majority     & Minority     & Majority    & Minority \\ \hline
    
1   & 1     & \cellcolor{pink!20}\textbf{1,384}  & 384  & \cellcolor{green!10}\textbf{0.91}  & 0.33      \\

2   & 1     & \cellcolor{pink!20}\textbf{1,263}  & 152  & \cellcolor{green!10}\textbf{0.97}  & 0.24      \\

3   & 2     & \cellcolor{pink!20}\textbf{2,001}  & 286  & \cellcolor{green!10}\textbf{0.92}  & 0.17      \\

4   & 2     & \cellcolor{pink!20}\textbf{1,236}  & 97   & \cellcolor{green!10}\textbf{0.98}  & 0.48      \\

5   & 1     & \cellcolor{pink!20}\textbf{1,160}  & 140  & \cellcolor{green!10}\textbf{0.95}  & 0.41      \\

6   & 0     & \cellcolor{pink!20}\textbf{934}   & 467  & \cellcolor{green!10}\textbf{0.84}  & 0.47      \\

7   & 2     & \cellcolor{pink!20}\textbf{948}   & 806  & \cellcolor{green!10}\textbf{0.83}  & 0.70      \\

\hlineB{2}
\end{tabular}
\label{table:preliminary}
\end{table}

\section{Motivation}

In this section, we first conduct an empirical study on the PubMed dataset \cite{sen2008dataset1} to investigate the impact of divergent neighboring information across clients on minority nodes when jointly training GNNs in FGL. The observation from this study is consistent with our example in Figure~\ref{fig:example} and motivates us to learn global structure proxies as favorable neighboring information. We then develop theoretical analysis to explain how aligning neighboring information across clients can benefit node classification tasks in FGL.

\subsection{Empirical Observations} \label{subsection: preliminary}
To better understand the divergent neighboring information across clients with its impact on the node classification task in FGL, we conduct preliminary experiments to compare the performance of federated node classification with MLP and GNNs as local models on the PubMed dataset \cite{sen2008dataset1}.
Following the data partition strategy in previous studies \cite{huang2023federated, zhang2021fedsage}, we synthesize the distributed graph data by splitting each dataset into multiple communities via the Louvain algorithm \cite{blondel2008louvain}. We retain seven communities with the largest number of nodes; each community is regarded as an entire graph in a client. 

Table~\ref{table:preliminary} shows the statistics of each client. According to Table~\ref{table:preliminary}, although one client may have the majority class different from another, the average node-level homophily of the majority class is consistently higher than that of the other classes for all the clients. For instance, the nodes in client 2 that do not belong to class 1 have only 24\% neighbors from the same class on average. It means that the minority nodes will absorb unfavorable neighboring information via GNNs and probably be classified incorrectly. 

\begin{figure}[t]
\setlength {\belowcaptionskip} {-0.3cm}
\centering
\includegraphics[width=\linewidth]{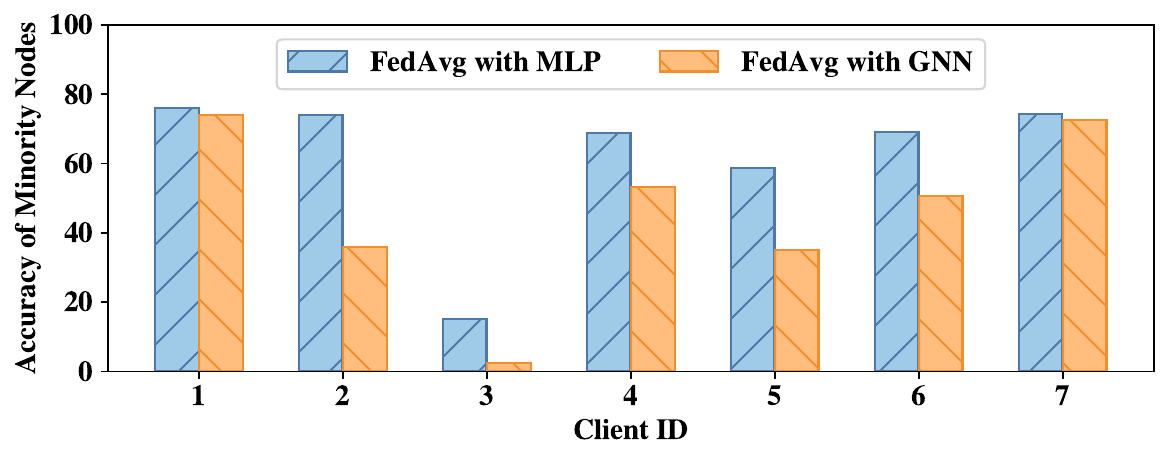}
\caption{Classification accuracy (\%) of minority nodes in each client by training MLP and GNN via FedAvg over the PubMed dataset. Average accuracy for all nodes: 82.35\% for MLP VS 87.06\% for GNN.}
\label{fig:preliminary}
\end{figure}

To validate our conjecture, we perform collaborative training for MLPs and GNNs following the standard FedAvg  \cite{mcmahan2017fl} over the PubMed dataset. Figure~\ref{fig:preliminary} illustrates the classification accuracy of minority nodes in each client by MLPs and GNNs. We can observe that MLPs consistently perform better than GNNs on minority nodes across the clients, although GNNs have higher overall accuracy for all nodes. Given that MLPs and GNNs are trained over the same node label distribution, we argue that the performance gap on minority nodes results from aggregating adverse neighboring information from other classes via the message-passing mechanism in GNNs, especially from the majority class. On the contrary, MLPs only need node features and do not require neighboring information throughout the training; therefore, they can avoid predicting more nodes as the majority class.

\begin{figure*}[!t]
\centering
\includegraphics[width=\linewidth]{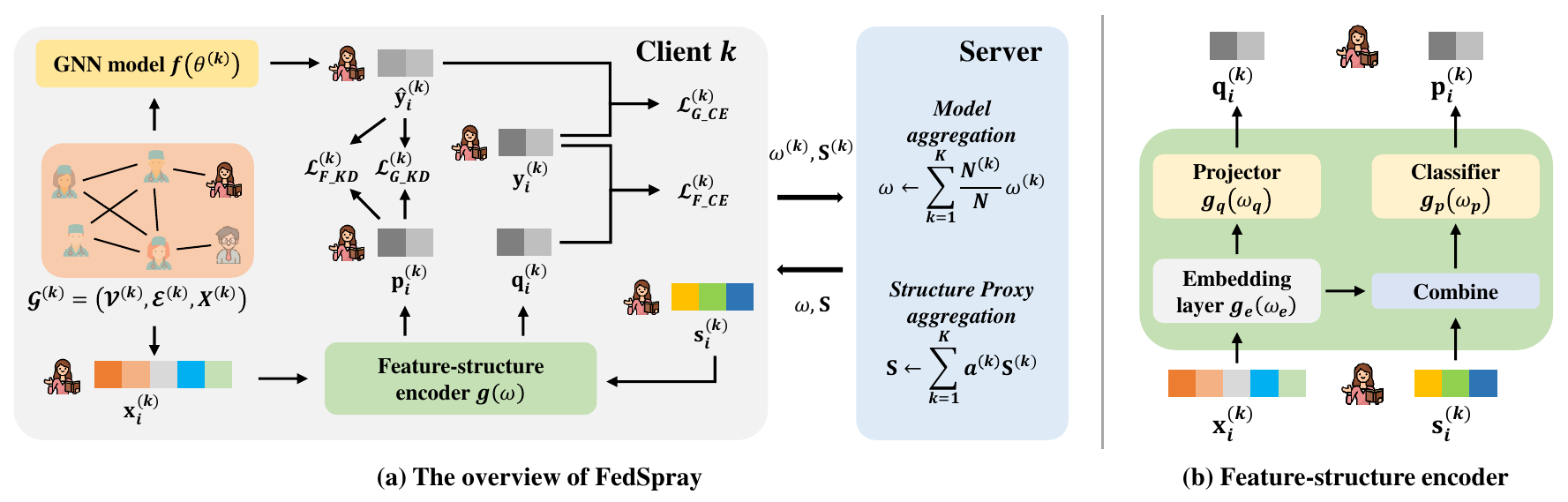}
\caption{ (a) An overview of the proposed FedSpray. The backbone of FedSpray is personalized GNN models $f(\theta^{(k)})$. A global feature-structure encoder $g(\omega)$ with structure proxies $\textbf{S}$ is also employed in FedSpray to tackle underrepresented node embeddings caused by adverse neighboring information in FGL. (b) An illustration of the feature-structure encoder in FedSpray.}
\label{fig:framework}
\end{figure*}

\subsection{Theoretical Motivation}
According to the above empirical observations, minority nodes with the original neighboring information are more likely to be misclassified. One straightforward approach to this issue is enabling nodes to leverage favorable neighboring information from other clients for generating node embeddings. Specifically, we consider constructing global neighboring information in the feature space. The server collects neighboring feature vectors from each client and computes the global class-wise neighboring information via FedAvg \cite{mcmahan2017fl}. We aim to theoretically investigate whether the global neighboring information can benefit node classification tasks when replacing the original neighbors of nodes. Following prevalent ways of graph modeling \cite{fortunato2016community,tsitsulin2022synthetic,ma2022homophily}, we first generate random graphs in each client using a variant of contextual stochastic block model \cite{tsitsulin2022synthetic} with two classes.

\subsubsection{\textbf{Random Graph Generation}}
The generative model generates a random graph in each client via the following strategy. In the generated graph $\mathcal{G}^{(k)}$ in client $k$, the nodes are labeled by two classes $c_1$ and $c_2$. For each node $v_i^{(k)}$, its initial feature vector $\textbf{x}_i^{(k)}\in \mathbb{R}^{d_x}$ is sampled from a Gaussian distribution $N(\boldsymbol{\mu}_1, \textbf{I})$ if labeled as class $c_1$ or $N(\boldsymbol{\mu}_2, \textbf{I})$ if labeled as class $c_2$ ($\boldsymbol{\mu}_1 \in \mathbb{R}^{d_x}$, $\boldsymbol{\mu}_2 \in \mathbb{R}^{d_x}$, and $\boldsymbol{\mu}_1 \neq \boldsymbol{\mu}_2$). For each client $k$, a neighbor of each node is from the majority with probability $p^{(k)}$ and from the minority with probability $1-p^{(k)}$. The ratio of minority nodes and majority nodes is $q^{(k)}$. In our setting, we assume $\frac{1}{2}<p^{(k)}<1$ and $0<q^{(k)}<1$. We denote each graph generated from the above strategy in client $k$ as $\mathcal{G}^{(k)}\sim\text{Gen}(\boldsymbol{\mu}_1, \boldsymbol{\mu}_2, p^{(k)}, q^{(k)})$.

\subsubsection{\textbf{Better Separability with Global Neighboring Information}} 
To figure out the influence of global neighboring information, we focus on the separability of the linear GNN classifiers with the largest margin when leveraging global neighboring information. Concretely, we aim to find the expected Euclidean distance from each class to the decision boundary of the optimal linear GNN classifier when it uses either the original neighboring information or the global neighboring information. We use $dist$ and $dist'$ to denote the expected Euclidean distances in these two scenarios, respectively. We summarize the results in the following proposition.

\begin{proposition} \label{proposition}
Given a set of $K$ clients, each client $k$ owns a local graph $\mathcal{G}^{(k)}\sim\text{Gen}(\boldsymbol{\mu}_1, \boldsymbol{\mu}_2, p^{(k)}, q^{(k)})$, $dist = \frac{||\boldsymbol{\mu}_1 - \boldsymbol{\mu}_2||_2}{2}$, which is smaller than $dist' = \left(1 + \sum_{k=1}^{K}(1-q^{(k)})(p^{(k)}-\frac{1}{2})\right)\frac{||\boldsymbol{\mu}_1 - \boldsymbol{\mu}_2||_2}{2}$.
\end{proposition}

A detailed proof can be found in Appendix~\ref{appendix: theoretical}. According to Proposition~\ref{proposition}, we will have a larger expected distance $dist'$ when using the global neighboring information. Typically, the larger the distance is, the smaller the misclassification probability is \cite{ma2022homophily}. Therefore, the optimal linear GNN classifier will obtain better classification performance. 

However, directly uploading neighboring feature vectors is implausible in FGL since it contains many sensitive raw features in the clients. To overcome this issue, we propose a novel framework FedSpray to learn global structure proxies in the latent space and elaborate on the details of FedSpray in Section~\ref{section: method}.

\section{Methodology} \label{section: method}
In this section, we present the proposed FedSpray in detail. Figure \ref{fig:framework}(a) illustrates an overview of FedSpray. The goal of FedSpray is to let the clients learn personalized GNN models over their private graph data while achieving higher performance by mitigating the impact of adverse neighboring information in GNN models. To reach this goal, FedSpray employs a lightweight global feature-structure encoder which learns class-wise structure proxies and aligns them on the central server. The feature-structure encoder generates reliable unbiased soft targets for nodes given their raw features and the aligned structure proxies to regularize local training of GNN models.

\subsection{Personalized GNN Model}
We first introduce personalized GNN models in FedSpray.

\subsubsection{\textbf{GNN backbone Model}}
Considering their exceptional ability to model graph data, we use GNNs as the backbone of the proposed framework. In this study, we propose to learn GNN models for each client in a personalized manner to tackle the data heterogeneity issue in FGL. Specifically, the personalized GNN model $f(\theta^{(k)})$ in client $k$ outputs the predicted label distribution $\hat{\textbf{y}}_i^{(k)}$ for each node $v_i^{(k)} \in \mathcal{V}_L^{(k)}$. Note that FedSpray is flexible. Any GNNs that follow the message-passing mechanism as the structure of Eq. (\ref{gnn}) can be used as the backbone, such as GCN \cite{kipf2016gcn} and SGC \cite{wu2019sgc}.

\subsubsection{\textbf{Loss formulation}}
During local training, $\theta^{(k)}$ can be updated by minimizing the cross-entropy loss between $\textbf{y}_i^{(k)}$ and $\hat{\textbf{y}}_i^{(k)}$ for each labeled node $v_i^{(k)} \in \mathcal{V}_L^{(k)}$
\begin{equation}
    \mathcal{L}_{G\_CE}^{(k)}
    = \frac{1}{|\mathcal{V}_L^{(k)}|}\sum_{v_i^{(k)} \in \mathcal{V}_L^{(k)}} \text{CE} (\textbf{y}_i^{(k)}, \hat{\textbf{y}}_i^{(k)}),
\end{equation}
where $\text{CE}(\cdot, \cdot)$ denotes the cross-entropy loss.
However, simply minimizing $\mathcal{L}_{G\_CE}^{(k)}$ can lead $\theta^{(k)}$ to overfitting during local training \cite{tan2022fedproto, li2021ditto}. In addition, the minority nodes are particularly prone to obtaining underrepresented embeddings due to biased neighboring information, as discussed above. 
To tackle this challenge, we propose to design an extra knowledge distillation term and use it to regularize local training of $\theta^{(k)}$. More concretely, we first employ the soft target $\textbf{p}_i^{(k)} \in \mathbb{R}^{d_p}$ for each node $v_i^{(k)} \in \mathcal{V}^{(k)}$ generated by the global feature-structure encoder to guide local training of $\theta^{(k)}$ in client $k$. Typically, we hope $\textbf{p}_i^{(k)}$ to be generated with unbiased neighboring information for node $v_i^{(k)}$ (we will elaborate on how to obtain proper $\textbf{p}_i^{(k)}$ in Section \ref{section:encoder}).
Then, we encourage $\hat{\textbf{y}}_i^{(k)}$ to approximate $\textbf{p}_i^{(k)}$ by minimizing the discrepancy between $\textbf{p}_i^{(k)}$ and $\hat{\textbf{y}}_i^{(k)}$ for each node $v_i^{(k)} \in \mathcal{V}^{(k)}$ in client $k$. Specifically, we achieve this via knowledge distillation \cite{hinton2015kd} as
\begin{equation} \label{g_kd}
    \mathcal{L}_{G\_KD}^{(k)}=\frac{1}{|\mathcal{V}^{(k)}|}\sum_{v_i^{(k)} \in \mathcal{V}^{(k)}} \text{KL} (\textbf{p}_i^{(k)}\Vert\hat{\textbf{y}}_i^{(k)}),
\end{equation}
where $\text{KL}(\cdot\Vert\cdot)$ is to compute the Kullback-Leibler divergence (KL-divergence).
Therefore, the overall loss for training $\theta^{(k)}$ in client $k$ can be formulated by combining the two formulations together
\begin{equation} \label{l_g}
   \mathcal{L}_G^{(k)} = \mathcal{L}_{G\_CE}^{(k)}+ \lambda_1\mathcal{L}_{G\_KD}^{(k)},
\end{equation}
where $\lambda_1$ is a predefined hyperparameter that controls the contribution of the knowledge distillation term in $\mathcal{L}_G^{(k)}$. When $\lambda_1$ is set as 0, FedSpray will be equivalent to training GNN models individually in each client. 

\subsection{Global Feature-Structure Encoder with Structure Proxies} \label{section:encoder}
In this subsection, we will elucidate our design for the global feature-structure encoder and class-wise structure proxies in FedSpray. The feature-structure encoder aims to generate a reliable soft target (i.e., $\textbf{p}_i^{(k)}$) for each node with its raw features and structure proxy. 

\subsubsection{\textbf{Structure Proxies}}
As discussed above, a minority node can obtain adverse neighboring information from its neighbors via the message-passing mechanism, given its neighbors are probably from other classes. To mitigate this issue, we propose to learn unbiased class-wise structure proxies in FedSpray, providing favorable neighboring information for each node. Here, we formulate each structure proxy in a vectorial form.
Let $\textbf{S} \in \mathbb{R}^{d_c \times d_s}$ denote class-wise structure proxies, and each row $\textbf{s}_j \in \textbf{S}$ denotes the $d_s$-dimensional structure proxy of the $j$-th node class. For each node $v_i^{(k)}\in \mathcal{V}^{(k)}_L$, its structure proxy $\textbf{s}_i^{(k)}$ will be $\textbf{s}_j$ if it is from the $j$-th class. Then, the structure proxies will be used as the input of the feature-structure encoder. 

\subsubsection{\textbf{Feature-Structure Encoder}}
In FedSpray, we employ a lightweight feature-structure encoder to generate a reliable soft target for a node with its raw feature and structure proxy as the input. Figure~\ref{fig:framework}(b) illustrates our design for the feature-structure encoder.
Let $g(\omega)$ denote the feature-structure encoder $g$ parameterized by $\omega$. Given a node $v_i^{(k)}\in \mathcal{V}^{(k)}_L$, the feature-structure encoder $g$ generates its soft target $\textbf{p}_i^{(k)}$ with its feature vector $\textbf{x}_i^{(k)}$ and structure proxy $\textbf{s}_i^{(k)}$ by
\begin{equation}
    \textbf{p}_i^{(k)}=g(\textbf{x}_i^{(k)}, \textbf{s}_i^{(k)};\omega).
\end{equation}

\noindent \textbf{Fusion of node features and structure proxies.}
Here, the problem is to determine a proper scheme for fusing a node's raw feature and its structure proxy in the feature-structure encoder. A straightforward way is to combine $\textbf{x}_i^{(k)}$ and $\textbf{s}_i^{(k)}$ together as the input of the feature-structure encoder. Ideally, $\textbf{s}_i^{(k)}$ can serve as surrogate neighboring information of node $v_i^{(k)}$ in the feature space. In this case, it requires $\textbf{s}_i^{(k)}$ to have the same dimension as that of $\textbf{x}_i^{(k)}$. However, this brings us a new challenge: when $\textbf{x}_i^{(k)}$ is of high dimension in graph data (e.g., 500 for PubMed \cite{sen2008dataset1}), directly learning high-dimensional $\textbf{s}_i^{(k)}$ in the feature space will be intractable. 
Considering this, we propose to learn $\textbf{s}_i^{(k)}$ in the latent space instead. Specifically, we split the feature-structure encoder into an embedding layer $g_e(\omega_e)$ and a classifier $g_p(\omega_p)$. The embedding layer first maps the raw feature $\textbf{x}_i^{(k)}$ of a node $\textbf{v}_i^{(k)}\in \mathcal{V}^{(k)}$ into the latent space to obtain its low-dimensional feature embedding $\textbf{e}_i^{(k)}$. Then we combine the feature embedding $\textbf{e}_i^{(k)}$ and the structure proxy $\textbf{s}_i^{(k)}$ together as the input of the classifier to get the soft target $\textbf{p}_i^{(k)}$. Mathematically, we can formulate this procedure as
\begin{equation} \label{soft_label}
\textbf{p}_i^{(k)} = g(\textbf{x}_i^{(k)}, \textbf{s}_i^{(k)};\omega) = g_p(\text{Combine}(\textbf{e}_i^{(k)}, \textbf{s}_i^{(k)}); \omega_p)
\end{equation}
where $\textbf{e}_i^{(k)} = g_e(\textbf{x}_i^{(k)}; \omega_e)$. Here, $\text{Combine}(\cdot,\cdot)$ is the operation to combine $\textbf{e}_i^{(k)}$ and $\textbf{s}_i^{(k)}$ together (e.g., addition).

\noindent \textbf{Structure proxies for unlabeled nodes.}
The feature-structure encoder can generate soft targets only for labeled nodes by Eq. (\ref{soft_label}) because the structure proxy $\textbf{s}_i^{(k)}$ requires the ground-truth label information of node $\textbf{v}_i^{(k)}$. To better regularize local training of the GNN model, we need to obtain soft targets for unlabeled nodes and use them to compute $\mathcal{L}_{G\_KD}^{(k)}$ by Eq. (\ref{g_kd}).
To achieve this, we design a projector $g_q(\omega_q)$ in the feature-structure encoder. It has the same structure as the classifier $g_p$. The difference is that the projector $g_q$ generates soft targets only based on feature embeddings. Specifically, we can obtain the soft label $\textbf{q}_i^{(k)}$ for each node $\textbf{v}_i^{(k)}\in \mathcal{V}^{(k)} \setminus \mathcal{V}^{(k)}_L$ with its feature embedding $\textbf{e}_i^{(k)}$ by
\begin{equation}
\textbf{q}_i^{(k)} = g_q(\textbf{e}_i^{(k)}; \omega_q).
\end{equation}
Therefore, we obtain the structure proxy $\textbf{s}_i^{(k)} = \langle \textbf{q}_i^{(k)}, \textbf{S}\rangle$ by computing the product of $\textbf{q}_i^{(k)}$ and $\textbf{S}$ for each node $\textbf{v}_i^{(k)}\in \mathcal{V}^{(k)} \setminus \mathcal{V}^{(k)}_L$. Since $\textbf{q}_i^{(k)}$ is normalized by the softmax operation, the inner product can also be viewed as the weighted average of $\textbf{S}$.

\subsubsection{\textbf{Loss formulation}}
During local training, we aim to update $\omega = \{\omega_e, \omega_p, \omega_q\}$ and $\textbf{S}$ using both ground-truth labels and predictions from the GNN model. Specifically, we formulate the overall loss for training $\omega$ and $\textbf{S}$ in client $k$ as
\begin{equation} \label{l_f}
    \mathcal{L}_F^{(k)} = \mathcal{L}_{F\_CE}^{(k)}+ \lambda_2\mathcal{L}_{F\_KD}^{(k)},
\end{equation}
where $\lambda_2$ is a hyperparameter. Here $\mathcal{L}_{F\_CE}^{(k)}$ is the average cross-entropy loss between $\textbf{y}_i^{(k)}$ and $\textbf{q}_i^{(k)}$ for each node $v_i^{(k)} \in \mathcal{V}_L^{(k)}$
\begin{equation}
    \mathcal{L}_{F\_CE}^{(k)} 
    = \frac{1}{|\mathcal{V}_L^{(k)}|}\sum_{v_i^{(k)} \in \mathcal{V}_L^{(k)}} \text{CE} (\textbf{y}_i^{(k)}, \textbf{q}_i^{(k)}).
\end{equation}
$\mathcal{L}_{F\_KD}^{(k)}$ is the average KL-divergence between $\textbf{p}_i^{(k)}$ and $\hat{\textbf{y}}_i^{(k)}$ to encourage $\textbf{p}_i^{(k)}$ to approach $\hat{\textbf{y}}_i^{(k)}$ for each node $v_i^{(k)} \in \mathcal{V}_L^{(k)}$
\begin{equation}
    \mathcal{L}_{F\_KD}^{(k)}=\frac{1}{|\mathcal{V}_L^{(k)}|}\sum_{v_i^{(k)} \in \mathcal{V}_L^{(k)}} \text{KL} (\hat{\textbf{y}}_i^{(k)}\Vert\textbf{p}_i^{(k)}).
\end{equation}

\subsection{Server Update}
As stated above, FedSpray will learn the feature-structure encoder and the structure proxies globally. In this subsection, we present the global update in the central server for the feature-structure encoder and the structure proxies, respectively.

\subsubsection{\textbf{Update global feature-structure encoder}}
During each round $r$, the server performs weighted averaging of local feature-structure encoders following the standard FedAvg \cite{mcmahan2017fl} with each coefficient determined by the local node size
\begin{equation} \label{global_update1}
    \omega_r \leftarrow \sum_{k=1}^K\frac{N^{(k)}}{N}\omega_r^{(k)}.
\end{equation}

\subsubsection{\textbf{Structure proxy alignment}} 
Instead of using the local node size, we propose to assign higher weights to majority classes than minority classes for structure proxy alignment. More specifically, the server updates global structure proxy $\textbf{s}_{j, r} \in \textbf{S}_r$ during round $r$ by
\begin{equation} \label{global_update2}
    \textbf{s}_{j, r} \leftarrow \sum_{k=1}^K\frac{a_j^{(k)}}{a_j}\textbf{s}_{j, r}^{(k)},
\end{equation}
where $a_j^{(k)}$ is the ratio of nodes from the $j$-th class among $\mathcal{V}_L^{(k)}$ in client $k$ and ${a_j} = \sum_{k=1}^K a_j^{(k)}$.

\begin{algorithm}[!t]
    \caption{FedSpray}
    \label{alg:algorithm}
    \raggedright
    
    \textbf{Input}: initial personalized $\theta^{(k)}$ for each client $k$, global $\omega_0$ and $\textbf{S}_0$\\
    \begin{algorithmic} 
    \FOR{each round $r=1, \cdots, R$}
    \FOR{each client $k$ \textbf{in parallel}}
    \STATE $\omega_r^{(k)}, \textbf{S}_r^{(k)}\leftarrow$ LocalUpdate $(\omega_{r-1}, \textbf{S}_{r-1})$
    \ENDFOR
    \STATE Update $\omega_r$  by Eq. (\ref{global_update1})
    \STATE Update $\textbf{S}_r$ by Eq. (\ref{global_update2})
    \ENDFOR
    \end{algorithmic}
    \vspace{0.5em}
    
    \textbf{LocalUpdate}$(\omega_{r-1}, \textbf{S}_{r-1})$: \\
    \begin{algorithmic}[1]
    \STATE \textcolor{gray}{================== Phase 1 =====================}
    \STATE $\textbf{e}_i^{(k)} = g_e(\textbf{x}_i^{(k)}; \omega_{e,r-1})$ 
    \STATE $\textbf{q}_i^{(k)} = g_q(\textbf{e}_i^{(k)}; \omega_{q,r-1})$
    \STATE Compute local $\textbf{s}_i^{(k)}$ from $\textbf{S}_{r-1}$
    \STATE $\textbf{p}_i^{(k)} = g_p(\text{Combine}(\textbf{e}_i^{(k)}, \textbf{s}_i^{(k)}); \omega_{p,r-1})$
    \STATE $\theta_r^{(k)} = \theta_{r-1}$ 
    \FOR{$t=1, \cdots, E$}
    \STATE $\hat{\textbf{y}}_i^{(k)} = f(\textbf{x}_i^{(k)}, \mathcal{G}^{(k)}; \theta_r^{(k)})$
    \STATE Compute $\mathcal{L}_G^{(k)}$ by Eq. (\ref{l_g}) using $\textbf{p}_i^{(k)}$
    \STATE Update the local GNN model $\theta_r^{(k)} \leftarrow \theta_r^{(k)} - \eta_f \nabla \mathcal{L}_G^{(k)}$
    \ENDFOR 
    \STATE \textcolor{gray}{================== Phase 2 =====================}  
    \STATE $\omega_r^{(k)} = \omega_{r-1}$
    \STATE $\hat{\textbf{y}}_i^{(k)} = f(\textbf{x}_i^{(k)}, \mathcal{G}^{(k)}; \theta_r^{(k)})$
    \FOR{$t=1, \cdots, E$}
    \STATE $\textbf{e}_i^{(k)} = g_e(\textbf{x}_i^{(k)}; \omega_{e,r}^{(k)})$ 
    
    \STATE $\textbf{q}_i^{(k)} = g_q(\textbf{e}_i^{(k)}; \omega_{q,r}^{(k)})$ 
    \STATE $\textbf{p}_i^{(k)} = g_p(\text{Combine}(\textbf{e}_i^{(k)}, \textbf{s}_i^{(k)}); \omega_{p,r}^{(k)})$
    \STATE Compute $\mathcal{L}_F^{(k)}$ by Eq. (\ref{l_f}) using $\hat{\textbf{y}}_i^{(k)}$
    \STATE Update the local feature-structure encoder $\omega_r^{(k)} \leftarrow \omega_r^{(k)} - \eta_g \nabla \mathcal{L}_F^{(k)}$
    \STATE Update the local structure proxy \\ $\textbf{s}_i^{(k)} \leftarrow \textbf{s}_i^{(k)} - \eta_s \nabla \mathcal{L}_F^{(k)}$  
    \ENDFOR
    \STATE Update $\textbf{s}_j \in \textbf{S}_r^{(k)}$ by averaging $\textbf{s}_i^{(k)}$ of nodes from class $j$
    \RETURN{$\omega_r^{(k)}, \textbf{S}_r^{(k)}$}
    \end{algorithmic}
\end{algorithm}

\subsection{Overall Algorithm}
Algorithm \ref{alg:algorithm} shows the overall algorithm of the proposed FedSpray. During each round, each client performs local updates with two phases. In Phase 1, each client trains its personalized GNN models for $E$ epochs. We first compute $\textbf{p}_i^{(k)}$ for node $v_i^{(k)}$ by the global feature-structure encoder $g(\omega_{r-1})$ with its feature $\textbf{x}_i^{(k)}$ and corresponding structure proxy $\textbf{s}_i^{(k)}$ (line 5). Then $\textbf{p}_i^{(k)}$ is utilized to compute $\mathcal{L}_G^{(k)}$ (line 9) for training the GNN model (line 10). 
In Phase 2, the feature-structure encoder and structure proxies will be optimized for $E$ epochs. In client $k$, we first obtain $\hat{\textbf{y}}_i^{(k)}$ for node $v_i^{(k)}$ by the up-to-date GNN model (line 14). $\hat{\textbf{y}}_i^{(k)}$ for node $v_i^{(k)}$ will be used to compute $\mathcal{L}_F^{(k)}$ (line 19). Then we update $\omega_{r,t}^{(k)}$ and $\textbf{s}_i^{(k)}$ via gradient descent (line 20-21). At the end of each round, $\textbf{s}_j \in \textbf{S}_r^{(k)}$ will be updated by averaging $\textbf{s}_i^{(k)}$ of nodes from the $j$-th class (line 23). At the end of each round, the local feature-structure encoder and structure proxies will be sent to the central server for training in the next round (line 24). In the central server, FedSpray updates the feature-structure encoder by Eq. (\ref{global_update1}) and aligns structure proxies by Eq. (\ref{global_update2}).


\begin{table*}[t]
\caption{Classification accuracy (Average$\pm$Std) of FedSpray and other baselines on node classification over four datasets. \textit{Overall} and \textit{Minority} represent all nodes and minority nodes in the test sets, respectively.}
\label{table:main}
\setlength\tabcolsep{10pt}
\begin{tabular}{c|c|cc|cc|cc}
\hlineB{2}
\multirow{2}{*}{Dataset} & \multirow{2}{*}{Method}  & \multicolumn{2}{c|}{{\qquad GCN \qquad}}      & \multicolumn{2}{c|}{SGC}      & \multicolumn{2}{c}{GraphSAGE}\\\cline{3-8}
 &                                  & Overall               & Minority              & Overall           & Minority              & Overall           & Minority          \\ \hline
\multirow{6}{*}{PubMed} & Local     & $87.49 \pm 0.24$   & $51.00 \pm 1.20$    & $86.27 \pm 0.34$   & $40.83 \pm 0.93$    & $86.86 \pm 0.26$    & $48.66 \pm 1.14$     \\
                        & Fedavg    & $87.06 \pm 0.61$   & $55.77 \pm 0.90$    & $82.23 \pm 1.89$   & $56.21 \pm 1.38$    & $85.92 \pm 0.87$    & $61.35 \pm 2.76$     \\
                        & APFL      & $86.44 \pm 0.66$   & $49.25 \pm 2.07$    & $83.10 \pm 0.34$   & $28.25 \pm 5.69$    & $86.23 \pm 0.55$   & $45.41 \pm 1.50$     \\
                        & GCFL      & $86.74 \pm 0.69$   & $48.85 \pm 3.25$    & $71.81 \pm 8.22$   & $51.10 \pm 8.66$    & $85.35 \pm 0.46$    & $45.60 \pm 2.53$     \\
                        & FedStar   & $81.25 \pm 0.67$   & $12.01 \pm 2.89$    & $82.06 \pm 1.32$   & $19.24 \pm 7.71$    & $80.57 \pm 1.21$    & $7.78 \pm 7.01$     \\
                        & FedLit    & $57.95 \pm 3.82$   & $47.39 \pm 2.84$    & $84.82 \pm 0.72$   & $57.54 \pm 1.76$    & $70.73 \pm 7.61$   & $57.15 \pm 8.35$     \\
                        & \textbf{FedSpray}  & \cellcolor{pink!20}$\mathbf{87.71 \pm 0.65}$   & \cellcolor{cyan!10}$\mathbf{62.12 \pm 2.73}$    & \cellcolor{pink!20}$\mathbf{87.13 \pm 1.41}$   & \cellcolor{cyan!10}$\mathbf{59.23 \pm 1.25}$    & \cellcolor{pink!20}$\mathbf{87.02 \pm 1.01}$   & \cellcolor{cyan!10}$\mathbf{61.59 \pm 0.96}$    \\ \hline

\multirow{6}{*}{{WikiCS}}& Local    & $81.43 \pm 0.58$   & $41.36 \pm 1.78$    & $81.66 \pm 0.62$   & $40.02 \pm 1.41$    & $81.57 \pm 0.35$    & $42.93 \pm 1.85$    \\
                        & Fedavg    & $80.53 \pm 0.74$   & $35.48 \pm 2.52$    & $79.90 \pm 0.60$   & $34.24 \pm 1.17$    & $80.23 \pm 0.20$   & $47.60 \pm 0.99$     \\
                        & APFL      & $79.81 \pm 0.72$   & $38.33 \pm 5.43$    & $78.46 \pm 0.63$   & $27.40 \pm 1.52$    & $80.54 \pm 0.64$   & $42.16 \pm 1.21$     \\
                        & GCFL      & $75.79 \pm 1.56$   & $36.94 \pm 1.80$    & $74.85 \pm 1.65$   & $29.08 \pm 0.69$    & $73.34 \pm 3.24$    & $37.79 \pm 2.90$     \\
                        & FedStar   & $75.61 \pm 0.53$   & $16.72 \pm 3.23$    & $76.95 \pm 0.73$   & $21.90 \pm 2.82$    & $74.48 \pm 0.51$    & $10.66 \pm 1.71$     \\
                        & FedLit    & $49.08 \pm 4.98$   & $29.85 \pm 2.85$    & $56.18 \pm 9.39$   & $32.79 \pm 3.73$    & $60.37 \pm 4.33$    & $36.90 \pm 4.32$     \\
                        & \textbf{FedSpray}  & \cellcolor{pink!20}$\mathbf{81.51 \pm 0.45}$   & \cellcolor{cyan!10}$\mathbf{47.43 \pm 1.31}$   & \cellcolor{pink!20}$\mathbf{81.87 \pm 0.59}$    & \cellcolor{cyan!10}$\mathbf{46.60 \pm 0.00}$ & \cellcolor{pink!20}$\mathbf{81.93 \pm 0.30}$   & \cellcolor{cyan!10}$\mathbf{52.04 \pm 0.51}$  \\ \hline

\multirow{6}{*}{{Physics}} & Local  & $94.62 \pm 0.16$   & $72.75 \pm 0.73$    & $94.82 \pm 0.28$   & $76.24 \pm 9.07$    & $94.14 \pm 0.30$    & $69.50 \pm 0.97$     \\
                        & Fedavg    & $94.13 \pm 0.40$   & $66.45 \pm 2.28$    & $94.40 \pm 0.25$   & $66.58 \pm 1.01$    & $94.60 \pm 0.34$    & $74.27 \pm 0.95$     \\
                        & APFL      & $94.27 \pm 0.20$   & $72.83 \pm 3.73$    & $94.52 \pm 0.27$   & $69.27 \pm 1.67$    & $84.31 \pm 3.76$   & $38.65 \pm 7.33$     \\
                        & GCFL      & $88.97 \pm 2.61$   & $60.90 \pm 2.04$    & $94.02 \pm 0.29$   & $66.54 \pm 1.87$    & $80.71 \pm 3.91$    & $50.22 \pm 5.20$     \\
                        & FedStar   & $89.86 \pm 0.43$   & $33.44 \pm 3.27$    & $91.37 \pm 0.40$   & $45.27 \pm 4.73$    & $89.78 \pm 0.41$    & $32.91 \pm 3.52$     \\
                        & FedLit    & $85.11 \pm 2.58$   & $60.57 \pm 2.42$    & $87.57 \pm 1.47$   & $61.96 \pm 0.81$    & $86.68 \pm 0.27$    & $66.36 \pm 0.88$     \\
                        & \textbf{FedSpray}  & \cellcolor{pink!20}$\mathbf{95.59 \pm 0.24}$   & \cellcolor{cyan!10}$\mathbf{80.98 \pm 1.39}$    & \cellcolor{pink!20}$\mathbf{95.08 \pm 0.32}$   & \cellcolor{cyan!10}$\mathbf{82.43 \pm 1.62}$    & \cellcolor{pink!20}$\mathbf{94.73 \pm 0.37}$    & \cellcolor{cyan!10}$\mathbf{83.26 \pm 1.25}$     \\ \hline

\multirow{6}{*}{{Flickr}} & Local   & $43.18 \pm 0.55$   & $25.96 \pm 1.94$    & $46.82 \pm 0.93$   & $25.39 \pm 1.46$    & $49.72 \pm 0.85$    & $25.25 \pm 1.70$     \\
                        & Fedavg    & $44.53 \pm 1.36$   & $26.45 \pm 0.46$    & $47.03 \pm 1.39$   & $27.24 \pm 2.52$    & $47.51 \pm 1.40$    & $26.13 \pm 0.82$     \\
                        & APFL      & $32.27 \pm 3.58$   & $19.44 \pm 6.16$    & $46.93 \pm 0.50$   & $23.50 \pm 0.49$    & $34.59 \pm 2.83$   & $18.39 \pm 2.96$     \\
                        & GCFL      & $47.31 \pm 1.29$   & $19.71 \pm 2.20$    & $46.56 \pm 1.71$   & $26.48 \pm 3.60$    & $44.84 \pm 2.10$    & $16.76 \pm 2.45$     \\
                        & FedStar   & $47.73 \pm 0.85$   & $13.82 \pm 3.05$    & $48.45 \pm 0.58$   & $14.59 \pm 3.39$    & $46.36 \pm 1.04$    & $11.33 \pm 4.38$     \\
                        & FedLit    & $45.38 \pm 1.73$   & $23.62 \pm 8.74$    & $49.62 \pm 0.36$   & $24.46 \pm 0.81$    & $44.06 \pm 2.26$    & $18.12 \pm 2.30$    \\
                        & \textbf{FedSpray}  & \cellcolor{pink!20}$\mathbf{48.21 \pm 1.03}$   & \cellcolor{cyan!10}$\mathbf{29.72 \pm 0.75}$    & \cellcolor{pink!20}$\mathbf{50.07 \pm 0.75}$   & \cellcolor{cyan!10}$\mathbf{28.46 \pm 2.12}$    & \cellcolor{pink!20}$\mathbf{51.45 \pm 0.72}$    & \cellcolor{cyan!10}$\mathbf{27.52 \pm 0.42}$     \\  

\hlineB{2}
\end{tabular}
\end{table*}

\subsection{Discussion}
FedSpray exhibits superior advantages from various perspectives, including communication efficiency, privacy preservation, and computational cost. We provide an in-depth discussion about FedSpray's principal properties as follows.

\subsubsection{\textbf{Privacy Preservation}}
The proposed FedSpray uploads the parameters of local feature-structure encoders following the prevalent frameworks in FL \cite{li2020fedprox, li2021ditto, li2021moon}. Here, we mainly discuss the privacy concern about uploading local structure proxies first. In fact, structure proxies naturally protect data privacy. First, they are synthetic 1D vectors to provide high-quality neighboring information in the latent space. In other words, they do not possess any raw feature information. Second, they are generated by averaging the structure proxies from the same class, which is an irreversible operation. 
Moreover, we can employ various privacy-preserving techniques to further improve confidentiality.

\subsubsection{\textbf{Communication Efficiency}}
The proposed FedSpray requires clients to upload local feature-structure encoders and structure proxies. As we introduced above, the feature-structure encoder is a relatively lightweight model. As for structure proxies, their size is generally much smaller than that of model parameters given $d_s \ll d_x$. In addition, we can further reduce the number of uploaded parameters by setting smaller $d_s$.

\subsubsection{\textbf{Computational Cost}}
The additional computational cost in FedSpray is mainly on local updates for the feature-structure encoder and structure proxies. Compared with GNN models, the feature-structure encoder and structure proxies require fewer operations for updating parameters. Training GNN models is usually time-consuming since GNN models need to aggregate node information via the message-passing mechanism during the forward pass \cite{zhang2022glnn}. However, the feature-structure encoder only incorporates node features and structure proxies with fully connected layers to obtain soft targets. Therefore, the time complexity of local updates for the feature-structure encoder and structure proxies will be smaller than GNN models. 
Let $N$, $d_x$, and $E$ denote the number of nodes of the local graph in a client, the number of node features, and the number of edges, respectively. Considering a 2-layer GCN model with hidden size $d_h$, its computational complexity is approximately $O(N d_hd_x+E d_x)$. Similarly, we can conclude that the computational complexity of the feature-structure encoder with the $d_s$-dimensional structure proxy is about $O(N d_sd_x)$, apparently smaller than the GCN model when we set $d_h=d_s$. Therefore, the feature-structure encoder in FedSpray does not introduce significant extra computational costs compared with FedAvg using GCN.
Furthermore, setting a smaller $d_s$ can also reduce computation costs.

\section{Experiments}

In this section, we conduct empirical experiments to demonstrate the effectiveness of the proposed framework FedSpray and perform detailed analysis of FedSpray.

\subsection{Experiment Setup}
\subsubsection{\textbf{Datasets}}
We synthesize the distributed graph data based on four common real-world datasets from various domains, i.e., PubMed \cite{sen2008dataset1}, WikiCS \cite{mernyei2020wikics}, Coauthor Physics \cite{shchur2018pitfalls}, and Flickr \cite{zeng2019graphsaint}. We follow the strategy in Section \ref{subsection: preliminary} to simulate the distributed graph data and summarize the statistics and basic information about the datasets in Appendix~\ref{appendix: datasets}. We randomly select nodes in clients and let 40\% for training, 30\% for validation, and the remaining for testing. We report the average classification accuracy for all nodes and minority nodes over the clients for five random repetitions.

\subsubsection{\textbf{Baselines}}
We compare FedSpray with six baselines including (1) \textbf{Local} where each client train its GNN model individually; (2) \textbf{FedAvg} \cite{mcmahan2017fl}, the standard FL algorithm; (3) \textbf{APFL} \cite{deng2020apfl}, an adaptive approach in personalized FL; (4) \textbf{GCFL} \cite{xie2021gcfl}, (5) \textbf{FedStar} \cite{tan2023fedstar}, and (6) \textbf{FedLit} \cite{xie2023fedlit}, three state-of-the-art FGL methods. More details about the above baselines can be found in Appendix~\ref{appendix: baselines}.

\subsubsection{\textbf{Hyperparameter setting}} 
As stated previously, FedSpray is compatible with most existing GNN architectures. In the experiments, we adopt three representative ones as backbone models: GCN \cite{kipf2016gcn}, SGC \cite{wu2019sgc}, and GraphSAGE \cite{hamilton2017graphsage}. Each GNN model includes two layers with a hidden size of 64. The size of feature embeddings and structure proxies is also set as 64. Therefore, the feature-structure encoder has similar amounts of parameters with GNN models. Each component in the feature-structure encoder is implemented with one layer. We use an Adam optimizer \cite{kingma2015adam} with learning rates of 0.003 for the global feature-structure encoder and personalized GNN models, 0.02 for structure proxies. The two hyperparameters $\lambda_1$ and $\lambda_2$ are set as 5 and 1, respectively. We run all the methods for 300 rounds, and the local epoch is set as 5.

\begin{figure}[t]
\centering
\includegraphics[width=\linewidth]{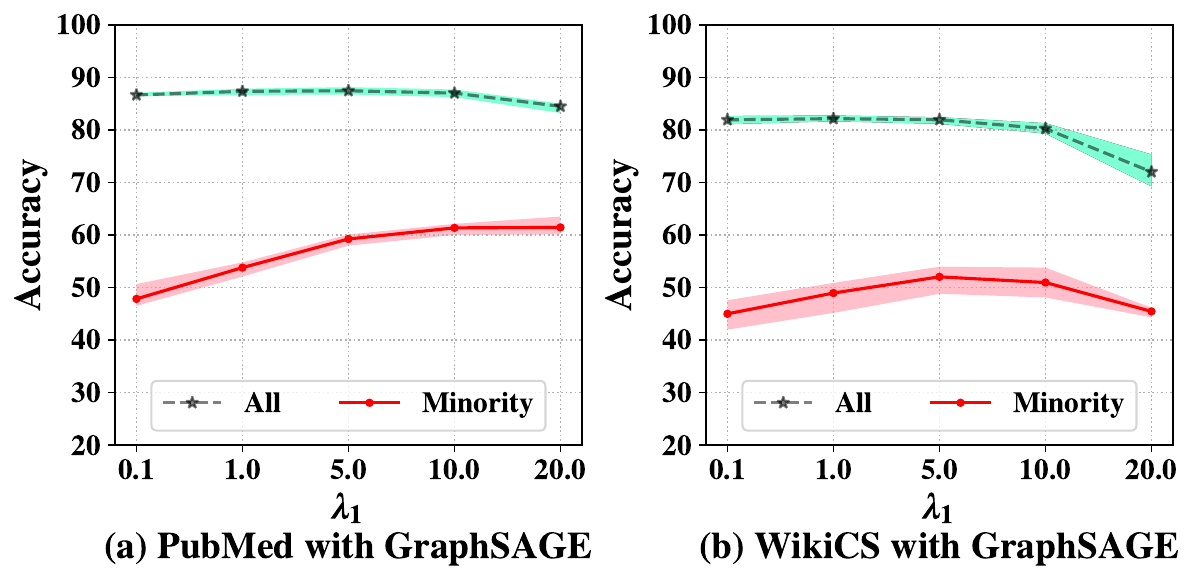}
\caption{Classification accuracy (\%) of FedSpray on all nodes and minority nodes in the test sets with different values of $\lambda_1$ over (a) PubMed and (b) WikiCS with GraphSAGE.}
\label{fig:lambda1}
\end{figure}

\subsection{Effectiveness of FedSpray}
We first show the performance of FedSpray and other baselines on node classification over the four datasets with three backbone GNN models. Table \ref{table:main} reports the average classification accuracy on all nodes and minority nodes in the test set across clients. 

First, we analyze the results of overall accuracy on all test nodes. According to Table \ref{table:main}, our FedSpray consistently outperforms all the baselines on node classification accuracy for overall test nodes across clients. Local and FedAvg achieve comparable performance over the four datasets. In the meantime, APFL does not surpass Local and FedAvg. As for FGL methods, GCFL, FedStar, FedLit fail to show remarkable performance gain. Although GCFL and FedStar tackle the data heterogeneity issue of graph structures across clients in FGL, they do not take the node-level heterophily into account. While FedLit models latent link types between nodes via multi-channel GNNs, it involves more GNN parameters that are hard to be well trained within limited communication rounds.

Second, we analyze the results of accuracy on minority nodes in the test set. Note that FedSpray aims to learn reliable unbiased structure information for guiding local training of personalized GNN models, particularly for minority nodes. We can observe that FedSpray outperforms all the baselines by a notable margin. Even though Local and FedAvg achieve comparable performance on overall test nodes, they show different accuracy results on minority nodes. Among the three FGL methods, FedStar encounters significant performance degradation on minority nodes since the design of structure embeddings in FedStar does not provide beneficial neighboring information for node-level tasks.

\begin{figure}[t]
\centering
\includegraphics[width=\linewidth]{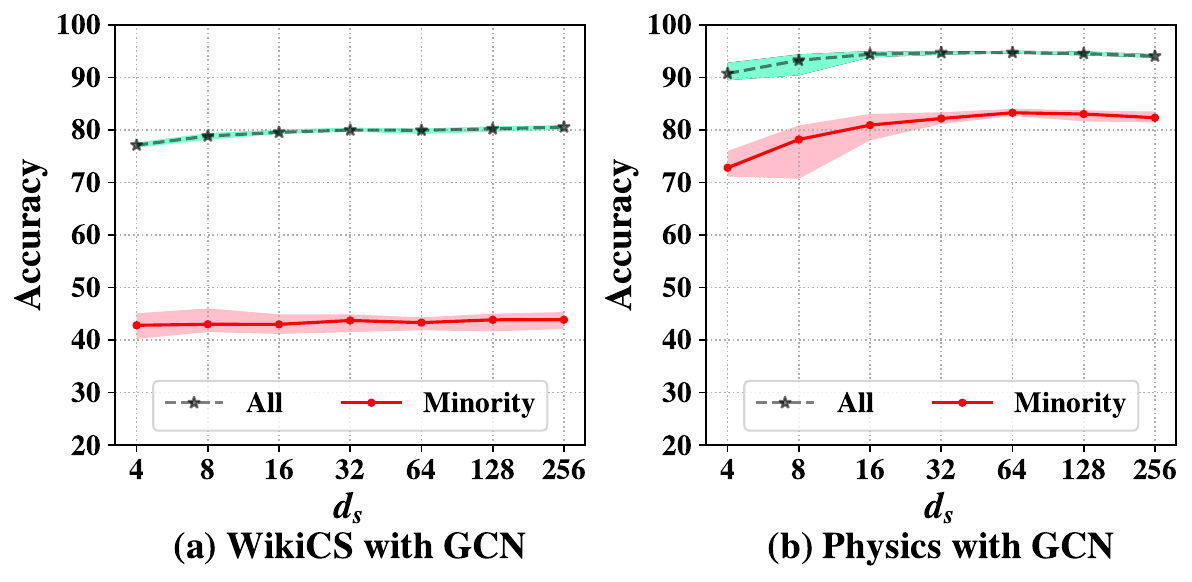}
\caption{Classification accuracy (\%) of FedSpray on all nodes and minority nodes in the test sets with different $d_s$ over (a) WikiCS and (b) Physics with GCN.}
\label{fig:size}
\end{figure}

\subsection{Analysis of FedSpray}

\subsubsection{\textbf{Influence of hyperparameter $\lambda_1$}}
The hyperparameter $\lambda_1$ controls the contribution of the regularization term in $\mathcal{L}_G{(k)}$. We conduct the sensitivity analysis on $\lambda_1$ in FedSpray. 
Figure \ref{fig:lambda1} reports the classification accuracy of FedSpray on all nodes and test nodes in the test sets with different values of $\lambda_1$ over PubMed (left) and WikiCS (right) with GraphSAGE. The accuracy on all nodes remains high when $\lambda_1$ is relatively small (i.e., $\lambda_1=0.1, 1, 5$). However, the accuracy of minority nodes will decrease when $\lambda_1$ is too small because the feature-structure encoder cannot sufficiently regularize local training of GNN models with too small $\lambda_1$. When $\lambda_1$ gets too large, the accuracy of all nodes decreases in both figures. In this case, the regularization term weighs overwhelmingly in the loss for training GNN models; then GNN models cannot be sufficiently trained with label information. According to the above observations, we will recommend 10 for PubMed with GraphSAGE and 5 for WikiCS with GraphSAGE as the best setting for $\lambda_1$.

\subsubsection{\textbf{Influence to structure proxy dimension}}
Since FedSpray incorporates structure proxies in the feature-structure encoder, we may set a different dimension $d_s$ of structure proxies. We evaluate the performance of FedSpray with different values of $d_s$ while fixing the hidden dimension of the GNN model as 64. Figure \ref{fig:size} demonstrates the classification accuracy of FedSpray on all nodes and test nodes in the test sets with different values of $d_s$ over WikiCS (left) and Physics (right) with GCN as the backbone. We can observe that FedSpray can obtain comparable accuracy with $d_s$ smaller than 64 (e.g., $d_s=32$). In the meantime, FedSpray does not obtain significant performance gain when $d_s$ is larger than 64. From the above observation, we can reduce communication and computation costs by setting $d_s$ a smaller value such as 32.

\begin{table}[t]
\caption{Classification accuracy (Average$\pm$Std) of FedSpray with $\textbf{S}=\textbf{0}$ over PubMed and Physics with GCN.}
\label{table:ablation}
\setlength\tabcolsep{7pt}
\begin{tabular}{cccc}
\hlineB{2}

Dataset                       & Method                  & Overall           & Minority \\ \hline
\multirow{2}{*}{PubMed} & FedSpray          & $\mathbf{87.71 \pm 0.65}$   & $\mathbf{62.12 \pm 2.73}$   \\
                        & FedSpray ($\textbf{S}=\textbf{0}$)  & $77.11 \pm 0.43$   & $42.01 \pm 0.81$   \\ \hline
\multirow{2}{*}{{Physics}} & FedSpray & $\mathbf{95.59 \pm 0.24}$   & $\mathbf{80.98 \pm 1.39}$   \\
                        & FedSpray ($\textbf{S}=\textbf{0}$)    & $93.23 \pm 0.27$   & $72.57 \pm 0.38$      \\

\hlineB{2}
\end{tabular}
\end{table}

\subsubsection{\textbf{Effectiveness of structure proxies.}}
In this study, we design structure proxies in FedSpray to serve as global unbiased neighboring information for guiding local training of GNN models. To validate the effectiveness of structure proxies, we investigate the performance of the proposed framework when structure proxies are removed. Specifically, we set class-wise structure proxies $\textbf{S}$ as $\textbf{0}$ consistently during training. We report the performance of FedSpray with $\textbf{S} = \textbf{0}$ over PubMed and WikiCS in Table~\ref{table:ablation}. According to Table~\ref{table:ablation}, we can observe that FedSpray suffers from significant performance degradation when removing structure proxies. It suggests that structure proxies play a significant role in FedSpray. Without them, the feature-structure encoder generates soft targets only based on node features \cite{zhang2022glnn}. In this case, the soft labels can be unreliable when node labels are not merely dependent on node features and, therefore, provide inappropriate guidance on local training of personalized GNN models in FedSpray.

\subsubsection{\textbf{More Experimental Results.}} Due to the page limit, we provide experimental results of FedSpray with varying local epochs in Appendix~\ref{appendix: results}.

\section{Related Work}
\subsection{Federated Learning}
Recent years have witnessed the booming of techniques in FL and its various applications in a wide range of domains, such as computer vision \cite{oh2022fedbabu,chen2022bridging},
healthcare \cite{liu2021feddg,sui2020feded}, 
and social recommendation \cite{liu2021fedsog,wu2021fedgnn}. The most important challenge in FL is data heterogeneity across clients (i.e., the non-IID problem). A growing number of studies have been proposed to mitigate the impact of data heterogeneity. 
For instance, FedProx \cite{li2020fedprox} adds a proximal term to the local training loss to keep the updated parameters close to the global model.
Moon \cite{li2021moon} uses a contrastive loss to increase the distance between the current and previous local models.
FedDecorr \cite{shi2022feddecorr} mitigates dimensional collapse to prevent representations from residing in a lower-dimensional space. In the meantime, a battery of studies proposed personalized model-based methods. 
For example, pFedHN \cite{shamsian2021pflhn} trains a central hypernetwork to output a unique personalized model for each client.
APFL \cite{deng2020apfl} learns a mixture of local and global models as the personalized model.
FedProto \cite{tan2022fedproto} and FedProc \cite{mu2023fedproc} utilize the prototypes to regularize local model training. 
FedBABU \cite{oh2022fedbabu} proposes to keep the global classifier unchanged during the feature representation learning and perform local adoption by fine-tuning in each client.

\subsection{Federated Graph Learning}
Due to the great prowess of FL, it is natural to apply FL to graph data and solve the data isolation issue. 
Recently, a cornucopia of studies has extended FL to graph data for different downstream tasks, such as node classification \cite{xie2023fedlit}, knowledge graph completion \cite{chen2021fede}, and graph classification \cite{xie2021gcfl,tan2023fedstar}, cross-client missing information completion \cite{zhang2021fedsage, peng2022fedni}. Compared with generic FL, node attributes and graph structures get entangled simultaneously in the data heterogeneity issue of FGL. To handle this issue, a handful of studies proposed their approaches. For example, GCFL \cite{xie2021gcfl} and FedStar \cite{tan2023fedstar} are two recent frameworks for graph classification in FGL. The authors of GCFL \cite{xie2021gcfl} investigate common and diverse properties in intra- and cross-domain graphs. They employ Clustered FL \cite{sattler2020cfl} in GCFL to encourage clients with similar properties to share model parameters. A following work FedStar \cite{tan2023fedstar} aims to jointly train a global structure encoder in the feature-structure decoupled GNN across clients. FedLit \cite{xie2023fedlit} mitigates the impact of link-type heterogeneity underlying homogeneous graphs in FGL via an EM-based clustering algorithm.

\section{Conclusion}
In this study, we investigate the problem of divergent neighboring information in FGL. With the high node heterophily, minority nodes in a client can aggregate adverse neighboring information in GNN models and obtain biased node embeddings. To grapple with this issue, we propose FedSpray, a novel FGL framework that aims to learn personalized GNN models for each client. FedSpray extracts and shares class-wise structure proxies learned by a global feature-structure encoder. The structure proxies serve as unbiased neighboring information to obtain soft targets generated by the feature-structure encoder. Then, FedSpray uses the soft labels to regularize local training of the GNN models and, therefore, eliminate the impact of adverse neighboring information on node embeddings. We conduct extensive experiments over four real-world datasets to validate the effectiveness of FedSpray. The experimental results demonstrate the superiority of our proposed FedSpray compared with the state-of-the-art baselines.

\begin{acks}
This work is supported in part by the National Science Foundation under grants IIS-2006844, IIS-2144209, IIS-2223769, IIS-2331315, CNS-2154962, and BCS-2228534; the Commonwealth Cyber Initiative Awards under grants VV-1Q23-007, HV-2Q23-003, and VV-1Q24-011; the JP Morgan Chase Faculty Research Award; the Cisco Faculty Research Award; and Snap gift funding.
\end{acks}

\bibliographystyle{ACM-Reference-Format}
\balance
\bibliography{reference}

\appendix

\newpage

\section{Proof of Proposition~\ref{proposition}} \label{appendix: theoretical}

\textsc{Proposition}~\ref{proposition}.
\textit{
Given a set of $K$ clients, each client $k$ owns a local graph $\mathcal{G}^{(k)}\sim\text{Gen}(\boldsymbol{\mu}_1, \boldsymbol{\mu}_2, p^{(k)}, q^{(k)})$, $dist = \frac{||\boldsymbol{\mu}_1 - \boldsymbol{\mu}_2||_2}{2}$, which is smaller than $dist' = \left(1 + \sum_{k=1}^{K}(1-q^{(k)})(p^{(k)}-\frac{1}{2})\right)\frac{||\boldsymbol{\mu}_1 - \boldsymbol{\mu}_2||_2}{2}$.
}

\begin{proof}
Without loss of generality, we assume that the majority class is $c_1$ for each client $k=1, 2, \cdots, M$ and $c_2$ for each client $k=M+1, M+2, \cdots, K$. Based on the neighborhood distributions, the neighboring features aggregated by the message-passing mechanism in GNNs follow Gaussian distribution
\begin{equation}
    \textbf{h}_i^{(k)} \sim N\left(p^{(k)}\boldsymbol{\mu}_1 + (1-p^{(k)}) \boldsymbol{\mu}_2, \frac{\textbf{I}}{\sqrt{|\mathcal{N}(v_i)|}}\right)
\end{equation}
for each client $k=1, 2, \cdots, M$, and 
\begin{equation}
    \textbf{h}_i^{(k)} \sim N\left((1-p^{(k)})\boldsymbol{\mu}_1 + p^{(k)} \boldsymbol{\mu}_2, \frac{\textbf{I}}{\sqrt{|\mathcal{N}(v_i)|}}\right)
\end{equation}
for each client $k=M+1, M+2, \cdots, K$.

The expectation of node embeddings after the message-passing mechanism will be $\mathbb{E}_{c_1}[\textbf{x}_i^{(k)}+\textbf{h}_i^{(k)}]$ for class $c_1$ and $\mathbb{E}_{c_2}[\textbf{x}_i^{(k)}+\textbf{h}_i^{(k)}]$ for class $c_2$. We omit the linear transformation because it can be absorbed in the linear GNN classifiers. The decision boundary of the optimal linear classifier is defined by the hyperplane $\mathcal{P}$ that is orthogonal to 
\begin{equation}
\begin{aligned}
    &\, \mathbb{E}_{c_1}[\textbf{x}_i^{(k)}+\textbf{h}_i^{(k)}] - \mathbb{E}_{c_2}[\textbf{x}_i^{(k)}+\textbf{h}_i^{(k)}] \\
    = & \, \mathbb{E}_{c_1}[\textbf{x}_i^{(k)}]+\mathbb{E}_{c_1}[\textbf{h}_i^{(k)}] - \mathbb{E}_{c_2}[\textbf{x}_i^{(k)}] - \mathbb{E}_{c_2}[\textbf{h}_i^{(k)}] \\
\end{aligned}
\end{equation}
For each client $k$, we have $\mathbb{E}_{c_1}[\textbf{h}_i^{(k)}]=\mathbb{E}_{c_2}[\textbf{h}_i^{(k)}]$. Therefore,
\begin{equation} \label{equation:p}
\begin{aligned}
    &\, \mathbb{E}_{c_1}[\textbf{x}_i^{(k)}+\textbf{h}_i^{(k)}] - \mathbb{E}_{c_2}[\textbf{x}_i^{(k)}+\textbf{h}_i^{(k)}] \\
    = & \, \mathbb{E}_{c_1}[\textbf{x}_i^{(k)}] - \mathbb{E}_{c_2}[\textbf{x}_i^{(k)}] = \boldsymbol{\mu}_1 - \boldsymbol{\mu}_2,
\end{aligned}
\end{equation}
and the distance from each class to $\mathcal{P}$ is 
\begin{equation}
    dist = \frac{||\boldsymbol{\mu}_1 - \boldsymbol{\mu}_2||_2}{2}.
\end{equation}

Let the server collect neighboring information from each client via FedAvg. The global neighboring information will be 
\begin{equation}
    \textbf{s}_1 = \sum_{k=1}^{M} \textbf{h}_i^{(k)}+\sum_{k=M+1}^{K} q^{(k)} \textbf{h}_i^{(k)}
\end{equation}
for class 1 and
\begin{equation}
    \textbf{s}_2 = \sum_{k=1}^{M} q^{(k)} \textbf{h}_i^{(k)}+\sum_{k=M+1}^{K} \textbf{h}_i^{(k)}
\end{equation}
for class 2. In this case, we replace $\textbf{h}_i^{(k)}$ in Eq. (\ref{equation:p}) and get the new hyperplane $\mathcal{P}'$ that is orthogonal to
\begin{equation}
\begin{aligned}
    &\mathbb{E}_{c_1}[\textbf{x}_i^{(k)}+\textbf{s}_1] - \mathbb{E}_{c_2}[\textbf{x}_i^{(k)}+\textbf{s}_2] \\
    = & \, \mathbb{E}_{c_1}[\textbf{x}_i^{(k)}]+\mathbb{E}_{c_1}[\textbf{s}_1] - \mathbb{E}_{c_2}[\textbf{x}_i^{(k)}] - \mathbb{E}_{c_2}[\textbf{s}_2] \\
    = & \, \mathbb{E}_{c_1}[\textbf{x}_i^{(k)}] - \mathbb{E}_{c_2}[\textbf{x}_i^{(k)}] + \mathbb{E}_{c_1}[\textbf{s}_1] - \mathbb{E}_{c_2}[\textbf{s}_2] \\
    = & \, \boldsymbol{\mu}_1 - \boldsymbol{\mu}_2+\mathbb{E}_{c_1}[\textbf{s}_1] - \mathbb{E}_{c_2}[\textbf{s}_2],
\end{aligned}
\end{equation}
where
\begin{align}
    & \mathbb{E}_{c_1}[\textbf{s}_1] - \mathbb{E}_{c_2}[\textbf{s}_2] \notag \\
    = & \, \mathbb{E}_{c_1}\left[\sum_{k=1}^{M} \textbf{h}_i^{(k)}+\sum_{k=M+1}^{K} q^{(k)} \textbf{h}_i^{(k)}\right] \notag \\
    &\quad\quad - \mathbb{E}_{c_2}\left[\sum_{k=1}^{M} q^{(k)} \textbf{h}_i^{(k)}+\sum_{k=M+1}^{K} \textbf{h}_i^{(k)}\right] \notag \\
    = & \sum_{k=1}^{M} \mathbb{E}_{c_1}[\textbf{h}_i^{(k)}]+\sum_{k=M+1}^{K} q^{(k)} \mathbb{E}_{c_1}[\textbf{h}_i^{(k)}] \notag \\
    &\quad\quad - \sum_{k=1}^{M} q^{(k)} \mathbb{E}_{c_2}[\textbf{h}_i^{(k)}]-\sum_{k=M+1}^{K} \mathbb{E}_{c_2}[\textbf{h}_i^{(k)}] \notag \\
    = & \sum_{k=1}^{M}(1-q^{(k)}) (p^{(k)} \boldsymbol{\mu}_1 + (1-p^{(k)}) \boldsymbol{\mu}_2) \notag \\
    &\quad\quad +\sum_{k=M+1}^{K} (q^{(k)}-1)((1-p^{(k)})\boldsymbol{\mu}_1 + p^{(k)} \boldsymbol{\mu}_2) \notag \\
    = & \left(\sum_{k=1}^{K}(1-q^{(k)})p^{(k)} - \sum_{k=M+1}^{K} (1-q^{(k)})\right)\boldsymbol{\mu}_1 \notag \\
    &\quad\quad + \left(\sum_{k=1}^{K}(q^{(k)}-1)p^{(k)} - \sum_{k=1}^{M} (q^{(k)}-1)\right)\boldsymbol{\mu}_2 \notag \\
    = & \sum_{k=1}^{K}(1-q^{(k)})p^{(k)}(\boldsymbol{\mu}_1 - \boldsymbol{\mu}_2)  \notag \\
    &\quad\quad + \sum_{k=1}^{M} (1-q^{(k)})\boldsymbol{\mu}_2 - \sum_{k=M+1}^{K} (1-q^{(k)})\boldsymbol{\mu}_1 \notag \\
    = & \sum_{k=1}^{K}(1-q^{(k)})(p^{(k)}-\frac{1}{2})(\boldsymbol{\mu}_1 - \boldsymbol{\mu}_2)  \notag \\
    &\quad\quad + (\frac{\boldsymbol{\mu}_1+\boldsymbol{\mu}_2}{2})\left(\sum_{k=1}^{M} (1-q^{(k)})- \sum_{k=M+1}^{K} (1-q^{(k)})\right). \notag 
\end{align}


Given the balanced global distribution where we have the same number of nodes from class $c_1$ and $c_2$, $\sum_{k=1}^{M}(1-q^{(k)})-\sum_{k=M+1}^{K}(1-q^{(k)})$ in the second term will be equal to 0. Therefore, the above equation can be simplified as

\begin{equation}
    \mathbb{E}_{c_1}[\textbf{s}_1] - \mathbb{E}_{c_2}[\textbf{s}_2] = \sum_{k=1}^{K}(1-q^{(k)})(p^{(k)}-\frac{1}{2})(\boldsymbol{\mu}_1 - \boldsymbol{\mu}_2).
\end{equation}
Then the new hyperplane $\mathcal{P}'$ is orthogonal to

\begin{equation}
    \left(1 + \sum_{k=1}^{K}(1-q^{(k)})(p^{(k)}-\frac{1}{2})\right)(\boldsymbol{\mu}_1 - \boldsymbol{\mu}_2),
\end{equation}
which is in the same direction of $\boldsymbol{\mu}_1 - \boldsymbol{\mu}_2$. Given $0<q^{(k)}<1$ and $\frac{1}{2}<p^{(k)}<1$ for each client $k$, the distance from each class to $\mathcal{P}'$ is 
\begin{equation}
    dist' = \left(1 + \sum_{k=1}^{K}(1-q^{(k)})(p^{(k)}-\frac{1}{2})\right)\frac{||\boldsymbol{\mu}_1 - \boldsymbol{\mu}_2||_2}{2},
\end{equation}
which completes the proof.

\end{proof}

\begin{table}[t]
\setlength\tabcolsep{5.8pt}
\centering
\caption{The statistics and basic information about the four datasets adopted for our experiments.}
\begin{tabular}{cccccc} \hlineB{2}
\textbf{Dataset}        & \textbf{PubMed}   & \textbf{WikiCS}   & \textbf{Physics}  & \textbf{Flickr}\\ \hline
\textbf{Clients}        & 7                 & 12                &   12              & 20  \\
\textbf{Node Features}  & 500               & 300               & 8,415             & 500  \\
\textbf{Average Nodes}     & 1,608             & 861               & 2,651             & 4,441 \\
\textbf{Average Edges}     & 3,600             & 11,721            & 14,790            & 14,331    \\
\textbf{Classes}        & 3                 & 10                & 5                 & 7       \\
\hlineB{2}
\end{tabular}
\label{table:dataset}

\end{table}

\begin{table}[t]
\caption{Classification accuracy of FedSpray and Fedavg on node classification over WikiCS with GCN.}
\label{table:epoch}
\setlength\tabcolsep{10pt}
\begin{tabular}{cccc}
\hlineB{2}
\textbf{Epochs}                  & \textbf{Node sets}     & \textbf{FedAvg}                & \textbf{FedSpray}                  \\ \hline     
\multirow{2}{*}{$E=3$}  & Overall       & $80.38 \pm 0.85$      & $81.37 \pm 0.39$          \\
                        & Minority      & $35.04 \pm 2.79$      & $47.43 \pm 1.39$          \\ \hline
\multirow{2}{*}{$E=5$}  & Overall       & $80.53 \pm 0.74$      & $81.51 \pm 0.45$          \\
                        & Minority      & $35.48 \pm 2.52$      & $47.43 \pm 1.31$          \\ \hline
\multirow{2}{*}{$E=10$} & Overall       & $80.23 \pm 0.70$      & $81.43 \pm 0.51$          \\
                        & Minority      & $35.13 \pm 2.47$      & $46.85 \pm 1.18$          \\ 
\hlineB{2}

\end{tabular}
\end{table}

\section{Experiment Details} \label{appendix: details}

\subsection{Datasets} \label{appendix: datasets}
Here we provide a detailed description of the four datasets we adopted to support our argument. These datasets are commonly used in graph learning from various domains: PubMed in citation network, WikiCS in web knowledge, Physics in co-author graph, and Flickr in social images. 
Table \ref{table:dataset} summarizes the statistics and basic information of the distributed graph data.

\subsection{Baselines} \label{appendix: baselines}
We compare our FedSpray with six baselines in our experiments. We provide the details of these baselines as follows.

\begin{itemize}
    \item \textbf{Local}: Models are locally trained on each client using its local data, without any communication with the server or other clients for collaborative training.

    \item \textbf{FedAvg} \cite{mcmahan2017fl}: It is a foundation method of FL that operates by aggregating local updates from clients and computing a weighted average of the updates to update the global model.

    \item \textbf{APFL} \cite{deng2020apfl}: APFL empowers clients to utilize a combination of local and global models as their personalized model. Additionally, during training, APFL autonomously determines the optimal mixing parameter for each client, ensuring superior generalization performance, even in the absence of prior knowledge regarding the diversity among the data of different clients.

    \item \textbf{GCFL} \cite{xie2021gcfl}: GCFL employs a clustering mechanism based on gradient sequences to dynamically group local models using GNN gradients, effectively mitigating heterogeneity in both graph structures and features.

    \item \textbf{FedStar} \cite{tan2023fedstar}: FedStar is devised to extract and share structural information among graphs. It accomplishes this through the utilization of structure embeddings and an independent structure encoder, which is shared across clients while preserving personalized feature-based knowledge. 

    \item \textbf{FedLit} \cite{xie2023fedlit}: FedLit is an FL framework tailored for graphs with latent link-type heterogeneity. It employs a clustering algorithm to dynamically identify latent link types and utilizes multiple convolution channels to adapt message-passing according to these distinct link types.

\end{itemize}

\subsection{Extra Experimental Results} \label{appendix: results}

\subsubsection{\textbf{Results with Varying Local Epochs}}
In FL, clients usually perform multiple local training epochs before global aggregation to reduce communication costs. We show the results of FedSpray and FedAvg with varying local epochs in Table \ref{table:epoch}. The results demonstrate that FedSpray can consistently outperform FedAvg with different local epochs.



\end{document}